\documentclass[a4paper,twoside]{article}

\usepackage{epsfig}
\usepackage{subcaption}
\usepackage{calc}
\usepackage{amssymb}
\usepackage{amstext}
\usepackage{amsmath}
\usepackage{amsthm}
\usepackage{multicol}
\usepackage{pslatex}
\usepackage{apalike}
\usepackage{algorithm2e}
\usepackage[bottom]{footmisc}
\usepackage{url}
\usepackage{graphicx}
\usepackage{SCITEPRESS}     % Please add other packages that you may need BEFORE the SCITEPRESS.sty package.

\begin{document}

\title{Evolutionary Transfer Learning for Dragonchess}

\author{\authorname{Jim O'Connor, Annika Hoag, Sarah Goyette, Gary B. Parker}
\affiliation{Computer Science Department, Connecticut College, New London, United States}
\email{\{joconno2, ahoag, sgoyette, parker\}@conncoll.edu}
}


\abstract{Dragonchess, a three-dimensional chess variant introduced by Gary Gygax, presents unique strategic and computational challenges that make it an ideal environment for studying the transfer of artificial intelligence (AI) heuristics across domains. In this work, we introduce Dragonchess as a novel testbed for AI research and provide an open-source, Python-based game engine for community use. Our research investigates evolutionary transfer learning by adapting heuristic evaluation functions directly from Stockfish, a leading chess engine, and subsequently optimizing them using Covariance Matrix Adaptation Evolution Strategy (CMA-ES). Initial trials showed that direct heuristic transfers were inadequate due to Dragonchess’s distinct multi-layer structure and movement rules. However, evolutionary optimization significantly improved AI agent performance, resulting in superior gameplay demonstrated through empirical evaluation in a 50-round Swiss-style tournament. This research establishes the effectiveness of evolutionary methods in adapting heuristic knowledge to structurally complex, previously unexplored game domains.}

\onecolumn \maketitle \normalsize \setcounter{footnote}{0} \vfill

\section{\uppercase{Introduction}}
\label{sec:introduction}
Game playing has proven over time to be an effective testbed for advancements in the field of artificial intelligence (AI). From early heuristic-based approaches in chess and checkers to modern deep reinforcement learning strategies exemplified by systems like AlphaZero \cite{silver2017mastering}, the field has progressed significantly in adapting AI to complex strategic environments. However, the vast majority of research has focused on games with relatively consistent board structures, primarily two-dimensional layouts with well-defined and static piece movement rules \cite{hu2024games}. As AI techniques continue to evolve, there is an increasing ability to extend these methods to more intricate game domains that introduce new structural and strategic complexities.

Dragonchess, a game developed by Gary Gygax, presents an ideal case for studying AI adaptation in novel game environments. Unlike conventional chess, Dragonchess is played on a three-tiered board, significantly increasing the state and action space of the game \cite{DragonChess}. The game’s unique structure introduces multi-level interactions, distinct movement rules for different piece types, and additional strategic depth not present in traditional chess-like games. These factors make it a particularly challenging domain for AI research, as traditional search-based methods and heuristic evaluation functions struggle with the game’s heightened complexity.

In this work, we investigate evolutionary transfer learning in the context of Dragonchess by adapting heuristics from Stockfish, the leading open-source chess engine \cite{stockfish2025}. Stockfish heuristics, including piece-square tables, material values, and mobility-based evaluations, have been highly optimized for standard chess but perform poorly when directly transferred to Dragonchess due to its distinct board structure and movement rules. To address this, we employ Covariance Matrix Adaptation Evolution Strategy (CMA-ES) \cite{ostermeier1994derandomized} to evolve these heuristics, fine-tuning piece values and evaluation functions to better fit the Dragonchess domain.

Our experiments demonstrate that while direct transfer of Stockfish heuristics results in suboptimal play, evolutionary optimization significantly enhances the agent’s performance, surpassing standard minimax-based agents and validating the efficacy of evolutionary transfer learning in this novel setting. By integrating prior knowledge with adaptive optimization, we provide a compelling case for evolutionary computation as a viable pathway to adapting AI across structurally complex and underexplored game domains.

More broadly, this work extends the applicability of AI-driven game playing techniques to three-dimensional board games, underscoring the importance of heuristic adaptation in novel strategic settings. By demonstrating that evolutionary transfer learning can effectively bridge knowledge across disparate game domains, this research contributes to the growing body of work exploring generalizable AI approaches to strategic decision-making.

\section{\uppercase{Related Works}}
The study of AI in game playing contexts has historically been dominated by deterministic, fully observable domains such as chess, Go, and shogi, where heuristic evaluation functions and deep search methods have enabled significant advancements in computational play. Early chess engines such as Deep Blue \cite{campbell2002deep} demonstrated the efficacy of brute-force search when combined with carefully tuned heuristics, while later methods, such as Monte-Carlo Tree Search (MCTS), introduced probabilistic decision-making that proved highly effective in games like Go \cite{gelly2012grand}. More recent breakthroughs in deep reinforcement learning have shifted focus towards self-play and learned evaluation functions, exemplified by AlphaZero \cite{silver2017mastering2} and MuZero\cite{schrittwieser2020mastering}. These methods, while highly successful, require extensive computational resources and training time, making them impractical for many emerging domains including games with complex three-dimensional structures like Dragonchess \cite{silver2017mastering3}.

Transfer Learning is a concept first introduced by Bozinovski and Fulgosi as a method for transferring knowledge in neural networks \cite{bozinovski1976influence}. Over the following decades, research has expanded into various domains, including image recognition and speech processing, demonstrating success in a variety of contexts \cite{bozinovski2020reminder}. By the 1990s, multi-task learning and more sophisticated theoretical foundations emerged, further refining the mechanisms by which knowledge from one domain could be leveraged in another \cite{caruana1997multitask}. More recently, deep learning and large-scale pretraining have shown the efficacy of transfer learning across multiple applications \cite{iman2023review}\cite{zhuang2020comprehensive}. For instance, Braylan and Miikulainen applied transfer learning to the problem of training video game agents with limited data \cite{braylan2016object}. A related concept brought up by Snodgrass and Ontanon maps training data from a different, similar domain onto the target domain when there is not enough training data to train models within the target domain \cite{snodgrass2016approach}. Transfer learning has also been used successfully in video game level design, transferring knowledge about the style of one level into the style of another \cite{sarkar2022tile2tile}.

Transfer learning offers a promising alternative for accelerating AI development in novel game domains by enabling agents to leverage knowledge from previously learned tasks, facilitating adaptation to new environments with reduced computational overhead \cite{lu2015transfer}. Evolutionary transfer learning, in particular, has shown promise in optimizing AI agents for complex strategic tasks by evolving and fine-tuning existing heuristics \cite{hou2019evolutionary}. By adapting pre-existing knowledge rather than relying solely on an agent learning from scratch, evolutionary approaches allow for efficient parameter tuning and strategic refinement in novel problem spaces.

\section{\uppercase{Dragonchess}}

Dragonchess, introduced by Gary Gygax in Dragon magazine in 1985, is a three-dimensional chess variant designed to introduce additional strategic complexity beyond traditional chess \cite{DragonChess}. The game is played on three vertically stacked 12×8 boards, representing different terrain types: the top board symbolizes the sky, the middle board represents the land, and the bottom board corresponds to the underworld. Each board introduces unique movement constraints and interactions, requiring players to think beyond the conventional two-dimensional constraints of standard chess.

\begin{figure}[!ht]
  \centering
  \includegraphics[width=\columnwidth]{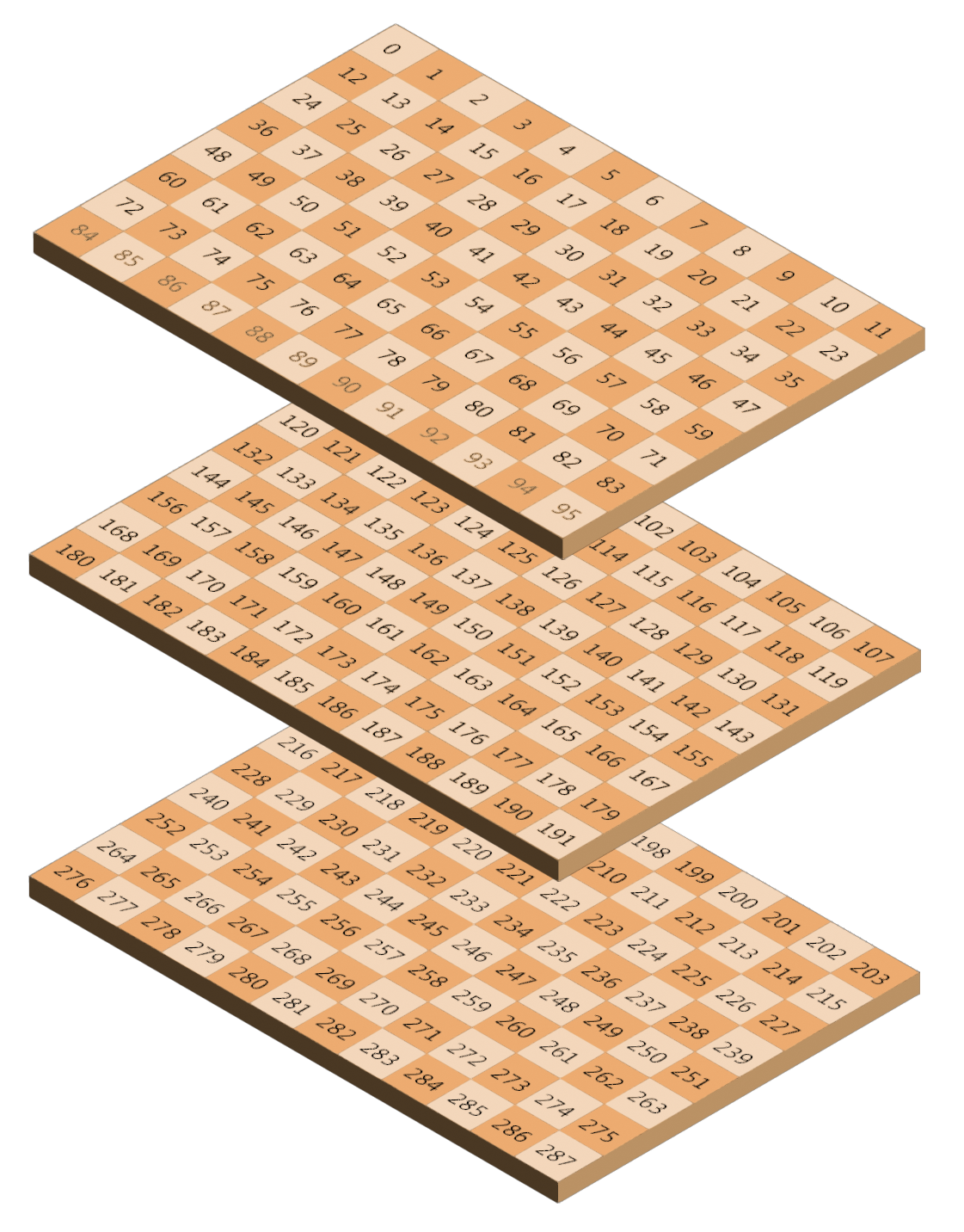}
  \caption{Each board layer (Sky, Ground, and Underground) consists of 8 rows and 12 columns, represented by integer indices from 0 to 287. Each array element at a given index stores an integer constant denoting piece type and ownership (positive integers for Gold, negative integers for Scarlet). This indexing approach enables efficient move generation and heuristic evaluations.}
  \label{fig1}
 \end{figure}

The game features a diverse set of pieces, many of which have asymmetric movement patterns and inter-board mobility. For example, the Dragon, a powerful piece, remains confined to the top board but possesses the unique ability to capture pieces on the middle board without moving. The Basilisk, located on the bottom board, can temporarily freeze enemy pieces on the middle board, adding a novel strategic dimension. Additionally, pieces such as the Griffin and the Hero can traverse multiple boards, reinforcing the importance of vertical positioning and inter-board interactions.

One of the primary challenges of Dragonchess lies in its significantly expanded state space. While standard chess has approximately $10^{43}$ possible positions, the three-tiered nature of Dragonchess increases the combinatorial complexity exponentially. The game’s additional layers of movement also introduce new strategic considerations, such as controlling vertical lanes to restrict opponent movement and coordinating multi-board attacks. These factors make heuristic evaluation more difficult, as conventional board evaluation functions struggle to capture the interplay between boards.

From an AI perspective, traditional search techniques such as minimax struggle with the sheer number of legal moves per turn, given that piece interactions occur across three layers rather than a single plane. Additionally, established heuristics for piece mobility and board control in chess do not directly transfer, as the relative value of a piece can vary significantly depending on its board location and potential mobility. For example, while a Mage (analogous to the Queen in chess) is highly mobile on the middle board, it becomes severely restricted when forced onto the top or bottom boards, requiring heuristic adjustments to account for these transitions.

These complexities make Dragonchess an ideal testbed for evaluating heuristic adaptation and evolutionary optimization techniques. The non-trivial transfer of chess heuristics to Dragonchess necessitates evolutionary fine-tuning to optimize piece values, movement priorities, and board control metrics. Understanding these challenges not only aids in developing stronger Dragonchess-playing AI but also provides broader insights into the adaptability of heuristic-based AI in novel strategic environments.

\section{\uppercase{Methodology}}

In this research, we develop an evolutionary knowledge transfer approach to AI agent optimization within the challenging game environment of Dragonchess. Our method consists of three primary phases: development of a novel Dragonchess game engine, direct heuristic knowledge transfer from Stockfish, and the evolutionary adaptation of these heuristics using the Covariance Matrix Adaptation Evolution Strategy (CMA-ES).

\begin{figure*}[!ht]
  \centering
  \includegraphics[width=\linewidth]{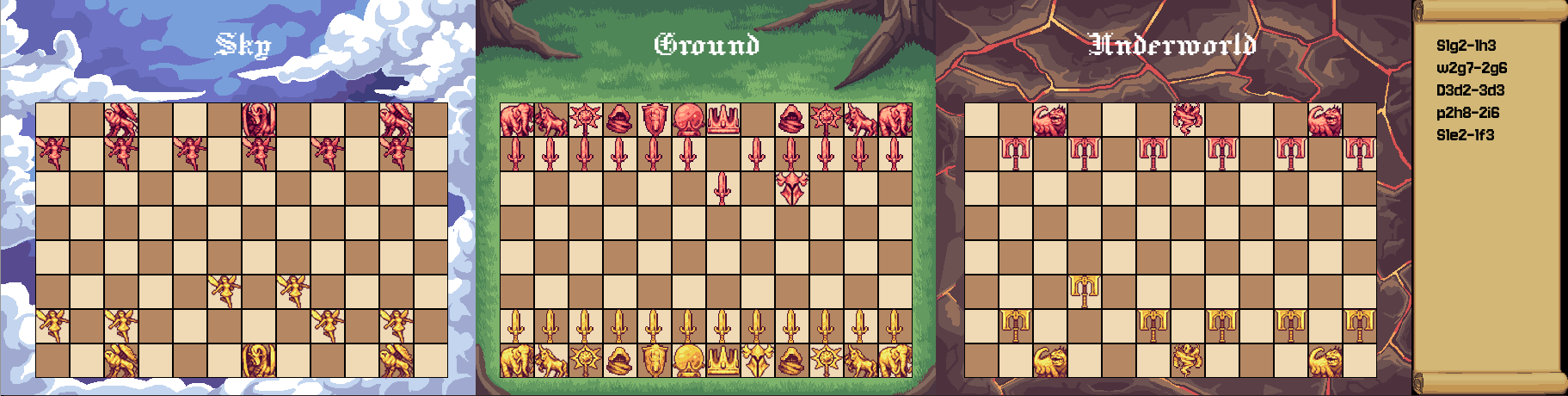}
  \caption{Screenshot from our Dragonchess engine's graphical interface built using PyGame. The visualization shows Dragonchess' characteristic three-layered board—Sky (top), Land (middle), and Underworld (bottom)—with distinct piece types and clear depiction of multi-layer interactions and movements unique to Dragonchess.}
  \label{fig2}
 \end{figure*}

\subsection{Dragonchess Game Engine Development}

To further support understanding of Dragonchess and facilitate broader AI research into the domain, we have developed an open-source Python-based Dragonchess engine. Figure~\ref{fig2} illustrates our implementation, highlighting the multi-layered board structure and clearly displaying piece positions and potential inter-board interactions. The graphical interface, built using the PyGame library, allows visual inspection of game states and moves, assisting both heuristic evaluation and qualitative analysis of AI performance.

Due to the complexity of Dragonchess, which involves three distinct boards of dimensions 12x8 stacked vertically, naive chess board representations would be inadequate for our purposes. To address this, we developed a novel representation specifically tailored for Dragonchess, in which each square on the three boards is assigned a unique integer index. These indices are used as direct positions within a NumPy array, where each element stores an integer constant representing the type of piece occupying that square. Positive integers indicate pieces belonging to the Gold player, while negative integers indicate Scarlet pieces.

This indexed integer representation allows efficient and rapid state evaluations, simplifies move generation, and enables straightforward position queries. By using integer-based indexing for each piece and board location, heuristic evaluations transferred from Stockfish can be computed efficiently, despite the increased complexity and dimensionality of Dragonchess. Representing each piece type distinctly within the indexed array enables streamlined calculations for threat detection, mobility assessment, and positional heuristics. The clarity and computational efficiency of this approach were critical in supporting the integration and evolutionary optimization of chess heuristics within the unique strategic environment presented by Dragonchess.

The engine manages the complexities associated with inter-board moves, piece-specific rules, and special abilities unique to Dragonchess. This includes logic for pieces like the Dragon and the Basilisk, whose moves involve interactions across multiple layers, capturing pieces from range, or temporarily disabling opponent pieces. Efficient move-generation and state-update algorithms were implemented to handle these complex dynamics without sacrificing computational performance.

\subsection{Transfer of Stockfish Heuristics}

The core innovation in our methodology involves transferring heuristic evaluation functions from Stockfish, a leading open-source chess engine, directly to the Dragonchess domain. Stockfish employs a suite of sophisticated heuristics, each finely tuned for classical chess, including piece valuations, positional heuristics, mobility metrics, threat detection, king safety evaluation, passed pawn assessments, material scoring, and imbalance corrections. These heuristics were transferred directly, without modification, to evaluate how standard chess heuristics perform within the structurally distinct and strategically intricate Dragonchess environment.

Specifically, the heuristics we implemented include:

\subsubsection{Material Values}

Stockfish assigns numeric valuations to chess pieces based on their strategic importance and average mobility within traditional chess. These standard valuations, derived by a community of chess players based on extensive empirical tuning, were transferred directly to analogous Dragonchess pieces to establish a baseline evaluation. The initial mappings were established based on the closest approximate role and mobility of Dragonchess pieces relative to traditional chess counterparts:

\begin{itemize}
    \item \textbf{Pawn (100 points)}: Applied to Dragonchess pieces fulfilling similar roles as low-value, front-line infantry, including the \textit{Sylph}, \textit{Warrior}, and \textit{Dwarf}. These pieces act as fundamental units, capturing and controlling space analogously to chess pawns.
    
    \item \textbf{Knight (320 points)}: Stockfish’s knight value was directly assigned to Dragonchess pieces with comparable strategic function, specifically the \textit{Unicorn} and \textit{Basilisk}. These pieces possess distinctive movement abilities, involving jumps or special inter-layer maneuvers reminiscent of knight moves.
    
    \item \textbf{Bishop (330 points)}: Assigned to Dragonchess pieces primarily valued for ranged control and diagonal mobility, such as the \textit{Cleric}, and the \textit{Mage}. These pieces mirror chess bishops by exerting long-range positional influence, though their movement and capture rules have notable domain-specific differences.
    
    \item \textbf{Rook (500 points)}: The rook’s valuation was applied to Dragonchess’s \textit{Hero}, \textit{Thief}, and \textit{Oliphant} pieces, reflecting their higher mobility, vertical traversal abilities, and strategic impact on board control similar to chess rooks.
    
    \item \textbf{Queen (900 points)}: The powerful \textit{Dragon} was assigned Stockfish’s queen valuation due to its dominating positional strength, significant mobility, and ability to control multiple squares from afar. This mapping acknowledges the Dragon’s high-impact potential analogous to the chess queen.
    
    \item \textbf{King (20,000 points)}: Dragonchess retains the singular strategic significance of the king, thus directly inheriting Stockfish’s substantial king valuation. This extremely high numeric value reflects the criticality of king safety and checkmate prevention, consistent with traditional chess heuristics.
\end{itemize}

These direct valuations provided a baseline for testing heuristic effectiveness in Dragonchess. However, due to substantial differences in piece mobility, capturing mechanics, and the multi-layered nature of Dragonchess, experiments showed that these direct valuations required significant evolutionary adjustment to achieve competitive gameplay.

\subsubsection{Piece-Square Tables (PSQT)}

Piece-square tables (PSQT) encode positional heuristics by assigning bonuses or penalties to pieces depending on their specific locations on the chessboard. Although Dragonchess differs significantly in its three-layered board structure and varying board sizes, we initially applied Stockfish's classical 8$\times$8 PSQT without adaptation. Each Dragonchess board was evaluated independently using the standard tables, despite potential mismatches in board geometry causing a lack of information on the sides of the Ground board and in total for the Sky and Underworld boards.

\subsubsection{Mobility Metrics}

Mobility heuristics quantify the strategic advantage gained from increased potential moves available to each piece. We directly incorporated Stockfish's mobility calculations into Dragonchess, considering the number of legal moves each piece could make. This metric was implemented separately for each piece type, and transferred without considering special Dragonchess rules, such as multi-level movement or board transitions.

\subsubsection{King Safety Evaluation}

The king safety heuristic estimates vulnerability based on proximity to enemy pieces, available escape routes, and defensive pawn structure in classical chess. We directly integrated this heuristic into Dragonchess, evaluating the king's exposure and threat-level across the three boards without adapting the rules to consider unique Dragonchess features such as vertical threats or remote attacks.

\subsubsection{Threat Detection}

Stockfish assesses threats based on potential captures and attacks, assigning penalties for positions where pieces are vulnerable. This heuristic was directly transferred by evaluating threatened squares and capturing potential across each individual Dragonchess board. We purposely did not account for inter-layer threats unique to Dragonchess to maintain the effect of the Stockfish oriented set of heuristics.

\subsubsection{Passed Pawns (Passed Pieces)}

In chess, passed pawn heuristics evaluate the strategic advantage of pawns unopposed in their advancement toward promotion. For Dragonchess, we transferred this concept directly as "passed pieces," evaluating positional bonuses for any pieces facing limited opposition in their direct line of advancement, despite differences in the promotion rules and board geometry of Dragonchess.

\subsubsection{Pawn Count}

Stockfish's heuristic calculates the number of pawns on the board, regardless of their position. The algorithm then provides a score calculated by the ratio of one player's pawns to another. This heuristic was conceptually transferred directly to Dragonchess, with the heuristic tracking the imbalance of Sylphs, Warriors, and Dwarves. These pieces were determined to be closest in both ability and strategic importance to the chess pawn.  

\subsubsection{Imbalance Total}

Stockfish incorporates heuristics that evaluate imbalances in material distributions, particularly the strategic consequences of uneven piece exchanges like bishop-pair advantages. This imbalance heuristic was transferred directly into Dragonchess, providing the strategic implications of asymmetric piece distributions without adaptation to Dragonchess-specific asymmetries or vertical interactions.

\subsubsection{Space Control}

The space heuristic quantifies the strategic advantage of controlling more squares, and consequently restricting opponent mobility. Directly transferred from Stockfish, our space heuristic simply counted empty squares controlled by each player, neglecting the vertical dimension and inter-board spatial interactions that distinguish Dragonchess.

\subsubsection{Board Activity Penalties}

Stockfish evaluates board activity by penalizing certain pieces occupying suboptimal or restricted positions. We incorporated these penalties directly into Dragonchess without modification, assessing penalties uniformly across boards regardless of layer-specific strategic nuances or special piece abilities unique to Dragonchess.

\subsubsection{Dragon Center Bonus (Adapted Queen Centralization)}

In classical chess, Stockfish assigns positional bonuses for centralized queens due to increased control. For Dragonchess, we adapted this heuristic to grant positional bonuses specifically for Dragons positioned centrally on the Sky board, directly transferring the centralization logic without extensive modification.

\subsubsection{Heuristic Total}

In this Stockfish-based assessment, all previous heuristics are added together to form a heuristic total. This value can be directly transferred to Dragonchess.

\subsection{Initial Transfer Performance}

Upon directly transferring these heuristics, our initial tests demonstrated that unmodified Stockfish heuristics performed poorly when directly applied to Dragonchess. The distinctive three-dimensional nature of the Dragonchess boards, coupled with unique piece movements and inter-board interactions, created critical mismatches in heuristic evaluations. Positional heuristics, particularly piece-square tables and threat evaluations, frequently underestimated or misjudged strategic opportunities and vulnerabilities. Mobility calculations and passed-piece heuristics also proved inadequate due to the unique vertical and inter-layer interactions present in Dragonchess. 

These initial observations underscored the need for adaptive heuristic optimization. Consequently, the transferred heuristics formed an effective baseline from which our evolutionary optimization process began. We then used CMA-ES to adapt these heuristics to better fit the complex and novel dynamics presented by Dragonchess.

\subsection{Evolutionary Optimization via CMA-ES}

To improve the effectiveness of these transferred heuristics, we employed an evolutionary optimization process using Covariance Matrix Adaptation Evolution Strategy (CMA-ES). CMA-ES is an evolutionary algorithm designed for the optimization of complex, high-dimensional, continuous parameter spaces, making it particularly suitable for adjusting heuristic parameters.

Our evolutionary strategy targeted a parameter vector comprising 25 numerical scaling factors. This vector encoded adjustments for piece values, positional heuristics, and other strategic factors originally derived from Stockfish. The specific parameterization was chosen based on preliminary analysis, which indicated sensitivity of Dragonchess heuristics to certain positional and material features.

Formally, we define our evolutionary objective as maximizing the performance of the agent:
\begin{equation}\label{eq1}
    \text{maximize}_{\,\mathbf{\theta}} \quad W(\theta)
\end{equation}

Where \( W(\theta) \) represents the expected win-rate of the Dragonfish agent with heuristic parameters \(\theta\), measured via win rate over simulated games. The parameter vector evolved by CMA-ES is represented as:
\begin{equation}\label{eq2}
    \theta = [w_1, w_2, w_3, \dots, w_{25}]
\end{equation}

These weights modify various heuristic components. Specifically, the first portion of weights (\(w_1\) through \(w_{14}\)) correspond are used to scale the piece valuations, while the remainder (\(w_{15}\) to \(w_{25}\)) are directly applied to the eleven previously listed dynamic positional evaluation components (mobility, threats, etc.). 

\begin{table*}[ht]
\caption{Results of the 50-Round Swiss-system Tournament}\label{tab:tournament_results} \centering
\begin{tabular}{|l|c|c|c|c|}
  \hline
    \textbf{Agent} & \textbf{Wins} & \textbf{Losses} & \textbf{Draws} & \textbf{Elo Rating} \\ \hline
    \textbf{Dragonfish (post-evolution)} & \textbf{32} & 17 & 1 & \textbf{1665.8} \\ \hline
    Jackman Minimax & 28 & 21 & 1 & 1604.7 \\ \hline
    Gygax Minimax & 28 & 22 & 0 & 1539.9 \\ \hline
    Dragonfish (pre-evolution) & 24 & 26 & 0 & 1488.6 \\ \hline
    Random Agent 1 & 21 & 23 & 6 & 1439.9 \\ \hline
    Random Agent 2 & 20 & 24 & 6 & 1449.7 \\ \hline
    Random Agent 3 & 20 & 26 & 4 & 1420.2 \\ \hline
    Random Agent 4 & 16 & 30 & 4 & 1337.2 \\ \hline
\end{tabular}
\end{table*}

\subsection{Evolutionary Optimization Procedure}

The evolutionary process involved the following iterative steps:

\begin{enumerate}
    \item Initialize a CMA-ES population from the direct-transfer heuristic parameters derived from Stockfish.
    \item Evaluate each candidate heuristic vector by simulating multiple Dragonchess games against standard baseline agents and recording win-loss-draw ratios.
    \item Select the highest-performing candidates based on their win rate and Elo ratings, updating the CMA-ES covariance matrix accordingly.
    \item Continue evolving heuristics over multiple generations, periodically evaluating against baseline agents to assess improvement and convergence.
\end{enumerate}

CMA-ES was chosen due to its robustness in handling high-dimensional continuous optimization problems, enabling efficient heuristic tuning without the need for gradient computations or explicit derivative information.

\section{Results}

To quantify the effectiveness of our evolved heuristics, we performed an empirical evaluation through a Swiss-style tournament involving several AI agents, each given exactly three seconds per move to maintain fair computational conditions. The tournament was conducted over 50 rounds, to provide statistically significant results. The agents evaluated included:
\begin{itemize}
    \item \textbf{Dragonfish (post-evolution)}: The agent employing evolved heuristics from CMA-ES.
    \item \textbf{Dragonfish (pre-evolution)}: Baseline agent utilizing unmodified Stockfish heuristics.
    \item \textbf{Jackman Minimax}: A standard minimax agent using piece weights recommended by Edward Jackman specifically for Dragonchess.
    \item \textbf{Gygax Minimax}: A baseline minimax agent using Gary Gygax’s original proposed weights.
    \item \textbf{Random Agents}: Standard agents that selected moves randomly, serving as a lower-performance benchmark.
\end{itemize}

To evaluate AI agent performance, we employed multiple complementary metrics. Tournament Elo ratings were calculated after each round using the standard Elo system, offering a quantifiable and standardized measure of relative agent strength. Win, loss, and draw ratios were recorded to directly assess performance improvements resulting from heuristic evolution.

The results of the 50-round Swiss-system tournament are summarized in Table 1. Elo ratings, which quantify relative player strength based on match outcomes, were calculated after each round to provide clear comparative analysis.

\section{\uppercase{Conclusions}}
\label{sec:conclusion}

In this paper, we introduced an evolutionary transfer learning approach to adapt heuristics from the well-established Stockfish chess engine to the complex, three-dimensional domain of Dragonchess. Our method demonstrated that, while direct transfer of chess heuristics provides a functional baseline, the unique strategic complexity and three-dimensional structure of Dragonchess require adaption or tuning to achieve competitive performance. By employing the Covariance Matrix Adaptation Evolution Strategy (CMA-ES), we successfully evolved the initially transferred heuristic values, significantly enhancing agent performance and achieving superior tournament results.

The evolved Dragonfish agent demonstrated the strongest performance in our empirical evaluation, attaining an Elo rating substantially higher than agents employing directly transferred heuristics or handcrafted heuristic values. Our results underscore the viability and effectiveness of evolutionary transfer learning, highlighting its potential as an effective approach for adapting existing heuristic knowledge across structurally distinct and strategically complex game domains.

By contributing both a novel domain and an accessible implementation, this research lays the groundwork for future investigations into evolutionary heuristic adaptation, knowledge transfer methods, and AI strategies for complex, multi-layered game environments. Future directions may include further refinement of heuristic functions, exploration of deep learning-based approaches, and comparative analyses using additional evolutionary optimization techniques within the Dragonchess environment.

\bibliographystyle{apalike}
{\small
\bibliography{references}}

\end{document}